\documentclass{article}
\usepackage{ijcai25}

\usepackage{times}
\usepackage{soul}
\usepackage{url}
\usepackage[hidelinks]{hyperref}
\usepackage[utf8]{inputenc}
\usepackage[small]{caption}
\usepackage{graphicx}
\usepackage{amsmath}
\usepackage{amsthm}
\usepackage{booktabs}
\usepackage{algorithm}
\usepackage{algorithmic}
\usepackage[switch]{lineno}
\usepackage{amssymb}
\usepackage{multirow}
\usepackage{multicol}

\usepackage{natbib}
\usepackage{subcaption}

\title{DC4CR: When Cloud Removal Meets Diffusion Control in Remote Sensing}

\author{
Zhenyu Yu$^2$
\and
Mohd Yamani Idna Idris$^2$
\and
Pei Wang$^{1, *}$\\
\affiliations
$^1$Kunming University of Science and Technology\\
$^2$Universiti Malaya\\
\emails
yuzhenyuyxl@foxmail.com, yamani@um.edu.my,
peiwang@kust.edu.cn
}

\begin{document}

\twocolumn[{
\renewcommand\twocolumn[1][]{#1}
\maketitle
\begin{center}
    \captionsetup{type=figure}
    \includegraphics[width=1.0\linewidth]{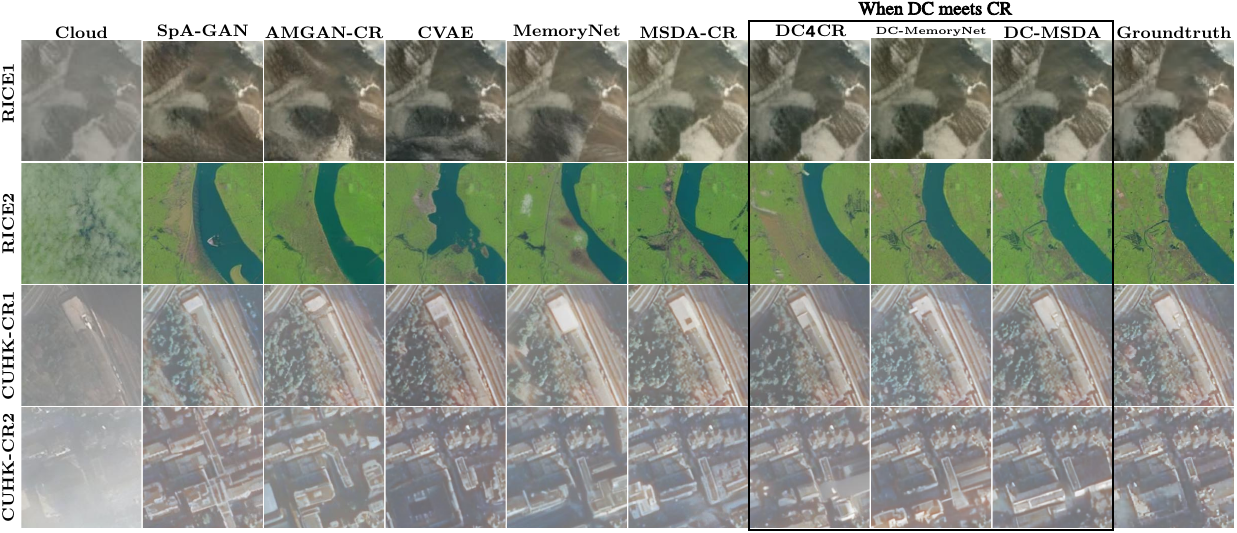} 
    \caption{Model comparison results. It indicates that our DC4CR model and other variants (DC-memorynet and DC-MSDA) excel across all conventional models. Note that these images can zoom in images, and the original images are shown in A.2.}
    \label{fig:model_compare}
\end{center}
}]

\begin{abstract}

Cloud occlusion significantly hinders remote sensing applications by obstructing surface information and complicating analysis. To address this, we propose \textbf{DC4CR} (Diffusion Control for Cloud Removal), a novel {multimodal diffusion-based} framework for cloud removal in remote sensing imagery. Our method introduces {prompt-driven control}, allowing selective removal of {thin and thick clouds} without relying on pre-generated cloud masks, thereby enhancing preprocessing efficiency and model adaptability. Additionally, we integrate {low-rank adaptation} for computational efficiency, {subject-driven generation} for improved generalization, and {grouped learning} to enhance performance on small datasets. Designed as a {plug-and-play module}, DC4CR seamlessly integrates into existing cloud removal models, providing a scalable and robust solution. Extensive experiments on the {RICE} and {CUHK-CR} datasets demonstrate {state-of-the-art performance}, achieving superior cloud removal across diverse conditions. This work presents a practical and efficient approach for remote sensing image processing with broad real-world applications.  

\end{abstract}

\section{Introduction}

High-resolution multispectral remote sensing has significantly advanced Earth observation, enabling precise analysis of land cover, vegetation health, urban development, and environmental changes \cite{yu2023impact,tan2023spatiotemporal}. However, cloud cover remains a major challenge in remote sensing image analysis, obstructing surface information and introducing uncertainty, particularly in multimodal datasets where different cloud formations increase restoration complexity~\citep{zi2018cloud, irons2012next,yang2020computer}. Thus, efficient cloud removal is a crucial preprocessing step for applications such as agriculture monitoring, disaster assessment, and land-use classification.

Cloud removal aims to reconstruct cloud-covered regions while preserving spatial and spectral consistency. 
\textbf{Physical model-based methods} rely on atmospheric correction and spectral analysis to detect and remove clouds~\citep{hu2015thin,yu2024capan,wang2024multi}. While effective for thin clouds, these methods struggle with complex cloud structures and varying atmospheric conditions, often failing to reconstruct occluded surface details~\citep{xu2019thin,yu2025yuan}. 
\textbf{Deep learning-based methods} have shown promising results by learning cloud removal patterns from large-scale datasets~\citep{pan2020cloud, xu2022attention,yu2025qrs}. Techniques such as convolutional neural networks (CNNs), generative adversarial networks (GANs), and variational autoencoders (VAEs) have been employed to enhance cloud removal accuracy. However, these models often lack adaptability to different cloud types and face challenges in generalizing to small dataset scenarios, limiting their robustness in real-world applications~\citep{ebel2022sen12ms,yu2025guideline}. 
\textbf{Multimodal approaches} integrate optical imagery with synthetic aperture radar (SAR) to exploit the complementary characteristics of different sensors~\citep{xu2019thin, pan2020cloud,yu2025ai}. These methods improve cloud removal performance by leveraging SAR's ability to penetrate clouds, but they require precise data alignment and registration, which remains a significant challenge for large-scale deployment~\citep{ebel2022sen12ms, xu2022attention}.

Recently, \textbf{diffusion models} have emerged as a powerful paradigm for image restoration due to their iterative refinement capabilities~\citep{ho2020denoising, yang2025pixel}, enabling the progressive reconstruction of missing or occluded regions through structured noise removal, making them well-suited for cloud removal tasks. However, directly applying diffusion models to cloud removal presents several challenges. First, adaptability to different cloud types remains a significant issue, as most existing methods employ a uniform restoration approach for both thin and thick clouds, often resulting in incomplete removal of thick clouds or excessive removal of thin clouds, thereby distorting surface details. Second, the high computational cost of diffusion models poses a practical limitation, as they require multiple denoising iterations, making them computationally expensive and impractical for large-scale remote sensing applications. Lastly, inconsistent color restoration after cloud removal is another major drawback, as existing methods often fail to preserve color consistency in reconstructed images, leading to mismatches or artifacts across different regions, ultimately affecting both visual quality and the performance of downstream tasks.

\textbf{Limitations of existing methods.} Despite advancements in cloud removal, current approaches still face fundamental challenges. One major limitation is the lack of a unified framework for different cloud types, as most methods treat thin and thick clouds separately, necessitating additional model training or preprocessing steps, thereby increasing complexity. Additionally, computational inefficiency remains a concern, particularly with diffusion models, which, despite their high restoration quality, require a multi-step denoising process that incurs substantial computational costs, making them difficult to scale for real-world applications. Furthermore, color inconsistencies after cloud removal often degrade visual quality and usability, as existing methods frequently fail to maintain color consistency between restored and surrounding areas, leading to mismatched tones that can negatively impact subsequent remote sensing analysis.

To address these challenges, we propose \textbf{D}iffusion \textbf{C}ontrol \textbf{for} \textbf{C}loud \textbf{R}emoval (\textbf{DC4CR}), a novel multimodal diffusion framework that introduces prompt-driven controllability for efficient and adaptive cloud removal. Our main \textbf{contributions} are:

\begin{itemize}
    \item \textbf{Prompt-based controllable cloud removal}. We introduce a {prompt-driven control} mechanism that enables a unified framework for removing both thin and thick clouds. Unlike prior methods that require separate models for different cloud types, our approach allows users to flexibly control cloud removal effects via prompt adjustments, enhancing adaptability across diverse cloud conditions.

    \item \textbf{Efficient diffusion-based cloud removal with Low-Rank Adaptation and Grouped Learning}. To reduce computational costs, we integrate {Low-Rank Adaptation (LoRA)} for efficient fine-tuning and {Grouped Learning} for progressive training. LoRA updates only a subset of parameters, while Grouped Learning enhances generalization by gradually increasing task complexity. These optimizations improve efficiency, enabling DC4CR to scale for large-scale remote sensing applications.

    \item \textbf{Image color optimization for enhanced consistency}. We integrate a VGG-19 module with Gram matrix constraints to improve color consistency. VGG-19 extracts style features for natural color restoration, while Gram matrices ensure uniform color distribution. This reduces color shifts and artifacts, enhancing the visual quality of remote sensing imagery.
\end{itemize}


\begin{figure*}[t] 
\centering
\includegraphics[width=0.99\textwidth]{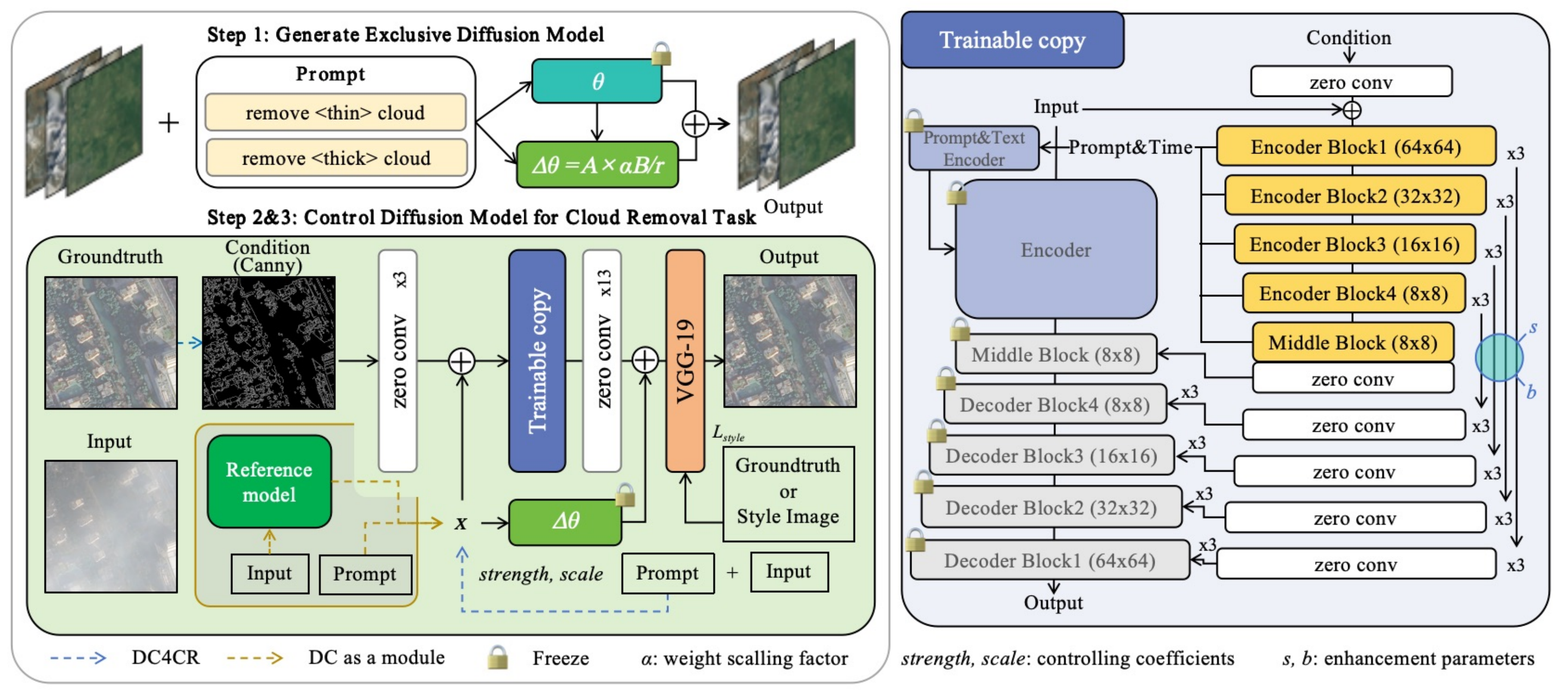} 
\caption{Overall workflow of the diffusion control for cloud removal (DC4CR) in remote sensing images. It illustrates the three main steps of the diffusion model designed for cloud removal tasks. \textbf{Step 1:} The text-to-image diffusion model is fine-tuned using prompts tailored to remote sensing characteristics to enhance the model's understanding and accuracy in generating remote sensing images. \textbf{Step 2 \& 3:} The cloud removal task is implemented by allowing users to select specific conditions to efficiently remove clouds and restore clear surface information. Meanwhile, optimize the quality of generated image.}
\label{fig:architecture}
\end{figure*}

\section{Related Work}

\subsection{Traditional Cloud Removal Methods}

\textbf{Multispectral methods} leverage wavelength-dependent absorption and reflection to reconstruct occluded landscapes \citep{xu2019thin}, but they struggle with thick clouds that completely block optical signals \citep{hu2015thin}.  
\textbf{Multitemporal methods} mitigate this issue by integrating cloud-free reference images captured at different times \citep{chen2019blind}. However, these methods heavily depend on the availability of clear-sky images, making them less effective in dynamic weather conditions.

\textbf{Other conventional approaches}, including interpolation \citep{ebel2022sen12ms}, wavelet transforms \citep{ji2020simultaneous}, and information cloning \citep{pan2020cloud}, have been explored to reconstruct cloud-contaminated regions. Additionally, sparse representation-based methods \citep{xu2022attention}, low-pass filtering \citep{rombach2022high}, and dictionary learning \citep{li2014recovering} have been investigated. 

Despite their effectiveness in certain scenarios, traditional methods generally fail to handle \textbf{diverse cloud types within a unified framework}. They often struggle with thick clouds, require manual intervention, and lack robustness in complex atmospheric conditions. These limitations have led to the adoption of deep learning-based approaches.

\subsection{Deep Learning-Based Cloud Removal}
Deep learning models have significantly improved CR performance by learning complex cloud removal patterns from large-scale datasets. 

\textbf{Generative models}, such as GANs and VAEs, have been widely adopted. SpA-GAN and AMGAN-CR employ spatial attention mechanisms to enhance cloud removal accuracy \citep{pan2020cloud, xu2022attention}, while CVAE introduces uncertainty modeling for improved robustness \citep{ding2022uncertainty}. However, these methods often struggle with \textbf{thick cloud occlusion}, leading to texture inconsistencies and loss of fine details.

\textbf{Memory-augmented architectures}, such as MemoryNet \citep{zhang2023memory} and MSDA-CR \citep{yu2022cloud}, integrate multiscale learning to improve structural consistency. Despite their effectiveness, these approaches require \textbf{significant computational resources}, making real-time processing challenging.

Furthermore, most deep learning-based methods \textbf{lack adaptability} to different cloud types and require explicit cloud masks, which introduce additional preprocessing complexity. To overcome these issues, more flexible and computationally efficient models are needed.

\subsection{Diffusion Models for Cloud Removal}
Diffusion models have emerged as powerful frameworks for image synthesis, video generation, and high-fidelity restoration \citep{saharia2022photorealistic, ho2022video, kawar2023imagic}. Unlike traditional deep learning-based CR models, diffusion models iteratively refine images through noise removal, making them particularly effective in reconstructing cloud-occluded regions \citep{ho2020denoising}.

In remote sensing, IDF-CR pioneered the application of diffusion models for CR by combining pixel-wise cloud reconstruction with iterative refinement \citep{wang2024idf}. Similarly, the DE method integrates reference priors to balance image fusion and employs a coarse-to-fine strategy for faster convergence \citep{sui2024diffusion}. However, existing diffusion-based methods remain \textbf{computationally expensive} due to their multiple denoising iterations, limiting their scalability for large-scale applications.

Additionally, most current diffusion models \textbf{apply uniform reconstruction strategies}, making it difficult to adapt to varying cloud types. Since cloud removal often introduces \textbf{color inconsistencies and artifacts}, a more controlled and adaptable diffusion process is required.

\begin{table*}[!ht]
    \caption{Comparison of cloud removal performance. Our model demonstrates superior results compared to other models.}
    \centering
    \resizebox{1.0\linewidth}{!}{
    \begin{tabular}{c|ccc|ccc|ccc|ccc}
    \hline
        \multirow{2}{*}{\textbf{Method}} & \multicolumn{3}{|c|}{\textbf{RICE1}} & \multicolumn{3}{|c|}{\textbf{RICE2}} & \multicolumn{3}{|c|}{\textbf{CUHK-CR1}} & \multicolumn{3}{|c}{\textbf{CUHK-CR2}} \\ \cline{2-13}
        & \textbf{PSNR}$\uparrow$ & \textbf{SSIM}$\uparrow$ & \textbf{LPIPS}$\downarrow$ & \textbf{PSNR}$\uparrow$ & \textbf{SSIM}$\uparrow$ & \textbf{LPIPS}$\downarrow$ & \textbf{PSNR}$\uparrow$ & \textbf{SSIM}$\uparrow$ & \textbf{LPIPS}$\downarrow$ & \textbf{PSNR}$\uparrow$ & \textbf{SSIM}$\uparrow$ & \textbf{LPIPS}$\downarrow$ \\ \midrule
        SpA-GAN & 28.509 & 0.9122 & 0.0503 & 28.783 & 0.7884 & 0.0963 & 20.999 & 0.5162 & 0.0830  & 19.680  & 0.3952  & 0.1201 \\ 
        AMGAN-CR & 26.497 & 0.912 & 0.0447 & 28.336 & 0.7819 & 0.1105 & 20.867 & 0.4986 & 0.1075  & 20.172  & 0.4900  & 0.0932 \\ 
        CVAE & 31.112 & 0.9733 & 0.0149 & 30.063 & 0.8474 & 0.0743 & 24.252 & 0.7252 & 0.1075  & 22.631  & 0.6302  & 0.0489 \\ 
        MemoryNet & 34.427 & 0.9865 & 0.0067 & 32.889 & 0.8801 & 0.0557 & 26.073 & 0.7741 & 0.0315  & 24.224  & 0.6838  & 0.0403 \\ 
        MSDA-CR & 34.569 & 0.9871 & 0.0073 & 33.166 & 0.8793 & 0.0463 & 25.435 & 0.7483 & 0.0374  & 23.755  & 0.6661  & 0.0433 \\ 
        DE-MemoryNet & \underline{35.525} & \underline{0.9882} & \underline{0.0055} & \underline{33.637} & 0.8825 & 0.0476 & \underline{26.183} & \underline{0.7746} & \underline{0.0290}  & \underline{24.348}  & \underline{0.6843}  & \underline{0.0369} \\ 
        DE-MSDA & 35.357 & 0.9878 & 0.0061 & 33.598 & \underline{0.8842} & \underline{0.0452} & 25.739 & 0.7592 & 0.0321  & 23.968  & 0.6737  & 0.0372 \\ 
        DC4CR & \textbf{35.542}  & \textbf{0.9887}  & \textbf{0.0054}  & \textbf{33.678}  & \textbf{0.8859}  & \textbf{0.0451}  & \textbf{26.291}  & \textbf{0.7790}  & \textbf{0.0285}  & \textbf{24.595}  & \textbf{0.6986}  & \textbf{0.0326}  \\ \bottomrule
    \end{tabular}
    }\label{table:comparision}
\end{table*}

\begin{table*}[t]
    \caption{Grouped training results. It includes the impact of ungrouped, 2-group, and 3-group training on model performance. }
    \centering
    \resizebox{1.0\linewidth}{!}{
    \begin{tabular}{c|ccc|ccc|ccc|ccc}
    \toprule
        \multirow{2}{*}{\textbf{Group}} & \multicolumn{3}{|c|}{\textbf{RICE1}} & \multicolumn{3}{|c|}{\textbf{RICE2}} & \multicolumn{3}{|c|}{\textbf{CUHK-CR1}} &  \multicolumn{3}{|c}{\textbf{CUHK-CR2}} \\ \cline{2-13}
        ~ & \textbf{PSNR}$\uparrow$ & \textbf{SSIM}$\uparrow$ & \textbf{LPIPS}$\downarrow$ & \textbf{PSNR}$\uparrow$ & \textbf{SSIM}$\uparrow$ & \textbf{LPIPS}$\downarrow$ & \textbf{PSNR}$\uparrow$ & \textbf{SSIM}$\downarrow$ & \textbf{LPIPS}$\downarrow$ & \textbf{PSNR}$\uparrow$ & \textbf{SSIM}$\uparrow$ & \textbf{LPIPS}$\downarrow$ \\ \midrule
        No grouping & 35.535  & \underline{0.9882}  & \underline{0.0056}  & 33.670  & \underline{0.8840}  & 0.0458  & 26.284  & 0.7736  & 0.0291  & \underline{24.594}  & 0.6911  & \underline{0.0331}  \\ 
        2 Groups & \underline{35.538}  & 0.9866  & 0.0062  & \underline{33.676}  & 0.8789  & \underline{0.0455}  & \underline{26.288}  & \underline{0.7783}  & \underline{0.0289}  & 24.586  & \underline{0.6936}  & \underline{0.0331}  \\ 
        3 Groups & \textbf{35.542}  & \textbf{0.9887}  & \textbf{0.0054}  & \textbf{33.678}  & \textbf{0.8859}  & \textbf{0.0451}  & \textbf{26.291}  & \textbf{0.7790}  & \textbf{0.0285}  & \textbf{24.595}  & \textbf{0.6986}  & \textbf{0.0326}  \\ \bottomrule
    \end{tabular}
    }\label{table:progressive}
\end{table*}

\section{Method}

\subsection{Model Overview}
This paper presents \textbf{DC4CR}, a diffusion-based framework for cloud removal in remote sensing imagery (see Figure \ref{fig:architecture}). The model consists of three main components:

\textit{Step 1}: \textbf{Text-to-Image Diffusion Model.} We fine-tune the Stable Diffusion model on remote sensing images using controlled prompts to ensure the generation of high-quality cloud-free imagery.

\textit{Step 2}: \textbf{Cloud Removal via Controlled Diffusion.} By leveraging diffusion control mechanisms, we enable flexible cloud removal under different conditions (e.g., cloud thickness and type), achieving precise surface reconstruction.

\textit{Step 3}: \textbf{Optimization and Modularization.} We incorporate FreeU to refine image quality and introduce a modular structure to enhance adaptability and generalization in complex remote sensing scenarios.

\subsection{Generating Remote Sensing Diffusion Model}

\textbf{Prompt Selection.}  
To enable adaptive cloud removal, we categorize clouds into two types: thin and thick. Prompt-based control is applied to condition the diffusion process using training directives such as:
\begin{center}
\texttt{remove <thin / thick> cloud}.
\end{center}
These prompts guide the model in recognizing different cloud characteristics and generating high-fidelity cloud-free outputs.




\textbf{Low-Rank Adaptation (LoRA).}  
To efficiently fine-tune the Stable Diffusion model for cloud removal, we employ LoRA \citep{hu2021lora}, which introduces low-rank decomposition to reduce computational cost while maintaining adaptation flexibility. The updated model parameters are defined as:
\begin{equation}
    \boldsymbol{\theta}_l = \boldsymbol{\theta} + \frac{\alpha}{r} \mathbf{A} \mathbf{B},
\end{equation}
where \( \boldsymbol{\theta} \) represents the original model parameters, \( \mathbf{A} \in \mathbb{R}^{m \times k} \) and \( \mathbf{B} \in \mathbb{R}^{k \times n} \) are low-rank matrices with \( k \ll \min(m, n) \), and \( \alpha \) is a scaling factor regulating adaptation strength. The term \( r \) represents the rank of the low-rank decomposition, controlling the dimensionality reduction of the weight update.

During training, only \( \mathbf{A} \) and \( \mathbf{B} \) are updated while keeping \( \boldsymbol{\theta} \) frozen, allowing efficient domain adaptation without overwriting pre-trained knowledge:
\begin{equation}
    \Delta \boldsymbol{\theta} = \frac{\alpha}{r} \mathbf{A} \mathbf{B}.
\end{equation}

By setting \( r \) appropriately, the model balances expressiveness and computational efficiency, reducing memory consumption while maintaining adaptation capability for small remote sensing datasets.

\textbf{Subject-Driven Generation.}  
To enhance model adaptability to diverse cloud structures, we employ DreamBooth \citep{ruiz2023dreambooth} for fine-tuning, using a combination of real and synthesized images to refine the diffusion process. Given a cloud-covered input image \( \mathbf{x} \), the model learns a domain-specific prior \( p(\mathbf{x}) \) conditioned on subject-specific embeddings \( \mathbf{z}_s \), ensuring accurate cloud removal while preserving structural details. The objective function is formulated as:
\begin{equation}
    \mathcal{L} = \lambda_1 \| \mathbf{x} - \mathbf{x}' \|^2 + \lambda_2 \cdot \text{SSIM}(\mathbf{x}, \mathbf{x}'),
\end{equation}
where \( \mathbf{x}' \) represents the generated cloud-free image, and \( \lambda_1, \lambda_2 \) are weighting factors balancing pixel-wise accuracy and perceptual similarity.

To improve image quality, we iteratively update the latent representation \( \mathbf{z}_s \) while keeping most pre-trained parameters frozen:
\begin{equation}
    \mathbf{z}_s^{(t+1)} = \mathbf{z}_s^{(t)} - \eta \nabla_{\mathbf{z}_s} \mathcal{L},
\end{equation}
where \( \eta \) is the learning rate. This process ensures high-fidelity reconstruction while maintaining structural and textural consistency.

\subsection{Cloud Removal with Controlled Diffusion}

\textbf{Conditional Control.}  
Following the principles of ControlNet \citep{zhang2023adding}, we introduce a control variable \( \mathbf{c} \) to guide the diffusion process. The denoising process follows an iterative refinement step:
\begin{equation}
    \mathbf{x}_{t+1} = \mathbf{x}_t - \gamma \nabla_{\mathbf{x}_t} \mathcal{L}_{\text{denoise}}(\mathbf{x}_t, \mathbf{c}),
\end{equation}
where \( \mathbf{x}_t \) represents the image at time step \( t \), \( \mathbf{c} \) is the control signal, \( \gamma \) is the step size, and \( \mathcal{L}_{\text{denoise}} \) incorporates both cloud removal constraints and structure preservation.


\textbf{Grouped Learning.}  
To improve model adaptability, we employ grouped learning, combining $k$-means clustering with progressive training. Given training samples \( \mathcal{D} = \{ (\mathbf{x}_i, \mathbf{y}_i) \}_{i=1}^{N} \), we compute a similarity metric \( S_i \) based on SSIM and MSE:

\begin{equation}
    S_i = \lambda_1 \cdot \text{MSE}(\mathbf{x}_i, \mathbf{y}_i) + \lambda_2 \cdot \text{SSIM}(\mathbf{x}_i, \mathbf{y}_i).
\end{equation}

Samples are clustered into three groups, \( \mathcal{D}_1, \mathcal{D}_2, \mathcal{D}_3 \), based on increasing complexity, where complexity is determined by the similarity score \( S_i \). 

The detail of progressive training:  
We employ a three-stage progressive training strategy to improve model adaptability. The dataset is clustered into three groups \( \mathcal{D}_1, \mathcal{D}_2, \mathcal{D}_3 \) based on increasing complexity. Training starts with simple cases in \( \mathcal{D}_1 \) to learn basic transformations, then fine-tunes on moderate samples in \( \mathcal{D}_2 \), and finally refines on the most challenging cases in \( \mathcal{D}_3 \). At each stage \( t \), the model updates as:

\begin{equation}
    \boldsymbol{\theta}_{t+1} = \boldsymbol{\theta}_t - \eta \nabla_{\boldsymbol{\theta}} \mathcal{L}(\mathcal{D}_t; \boldsymbol{\theta}_t).
\end{equation}

This gradual complexity increase stabilizes training, enhances generalization, and improves cloud removal accuracy.

\subsection{Optimizing Image Quality and Modularization}

\textbf{Feature Enhancement.}  
We integrate FreeU \citep{si2024freeu} to optimize feature representations by dynamically adjusting scaling and shifting parameters:
\begin{equation}
    \mathbf{h}_{\text{scaled}} = s_1 \cdot \mathbf{h} + s_2, \quad
    \mathbf{h}_{\text{shifted}} = b_1 \cdot \mathbf{h}_{\text{scaled}} + b_2,
\end{equation}
where \( \mathbf{h} \) is the feature map, and \( s_1, s_2, b_1, b_2 \) are learnable parameters. This approach reduces artifacts, ensuring more realistic textures.

\textbf{Image Color Optimization.}  
We utilize a VGG-19 module with Gram matrix constraints \citep{gatys2015texture} to enforce color consistency:
\begin{equation}
    \mathcal{L}_{\text{style}} = \sum_l w_l \| G_l^{\text{gen}} - G_l^{\text{ref}} \|^2,
\end{equation}
where \( G_l \) denotes Gram matrices computed from feature maps. This technique ensures harmonized color distributions across cloud-free images.

\textbf{Plug-and-Play Modular Design.}  
DC4CR adopts a modular design, allowing seamless integration into existing cloud removal frameworks. By incorporating LoRA, Grouped Learning, and Conditional Control, the model dynamically adjusts to diverse scenarios, ensuring both efficiency and accuracy.

\begin{algorithm}[ht]
\caption{DC4CR: Cloud Removal via Diffusion Model}
\label{alg:cloud_removal}
\begin{algorithmic}[1]
\REQUIRE Dataset $\mathcal{D} = \{ (\mathbf{x}_i, \mathbf{y}_i) \}_{i=1}^{N}$, Pre-trained model $\boldsymbol{\theta}_0$
\ENSURE Fine-tuned model $\boldsymbol{\theta}$
\STATE Initialize $\boldsymbol{\theta} \gets \boldsymbol{\theta}_0$, update LoRA $\mathbf{A}, \mathbf{B}$
\STATE Compute $S_i$ (SSIM + MSE), cluster into $\mathcal{D}_1, \mathcal{D}_2, \mathcal{D}_3$ via $k$-means
\FOR{$e = 1$ to $N_{\text{epochs}}$}
    \FOR{$\mathcal{D}_t \in \{\mathcal{D}_1, \mathcal{D}_2, \mathcal{D}_3\}$}
        \FOR{each $(\mathbf{x}_i, \mathbf{y}_i) \in \mathcal{D}_t$}
            \STATE Generate $\mathbf{x}_i' = f_{\boldsymbol{\theta}}(\mathbf{x}_i, p_i)$ with prompt $p_i$
            \STATE Apply controlled diffusion: $\mathbf{x}_{t+1} = \mathbf{x}_t - \gamma \nabla_{\mathbf{x}_t} \mathcal{L}_{\text{denoise}}(\mathbf{x}_t, \mathbf{c}_i)$
            \STATE Compute total loss: 
            \STATE \quad $\mathcal{L} = \lambda_1 \| \mathbf{x}_i - \mathbf{x}_i' \|^2 + \lambda_2 \text{SSIM}(\mathbf{x}_i, \mathbf{x}_i') + \lambda_3 \mathcal{L}_{\text{style}}(\mathbf{x}_i', \mathbf{y}_i)$
            \STATE Update latent embedding: $\mathbf{z}_s^{(t+1)} = \mathbf{z}_s^{(t)} - \eta \nabla_{\mathbf{z}_s} \mathcal{L}$
            \STATE Apply FreeU enhancement: $\mathbf{h}_{\text{shifted}} = b_1 (s_1 \mathbf{h} + s_2) + b_2$
            \STATE Update $\boldsymbol{\theta}$ using Adam optimizer
        \ENDFOR
    \ENDFOR
\ENDFOR
\STATE \textbf{Return} $\boldsymbol{\theta}$
\end{algorithmic}
\end{algorithm}

\begin{figure*}[ht] 
    \centering
    \includegraphics[width=1.0\linewidth]{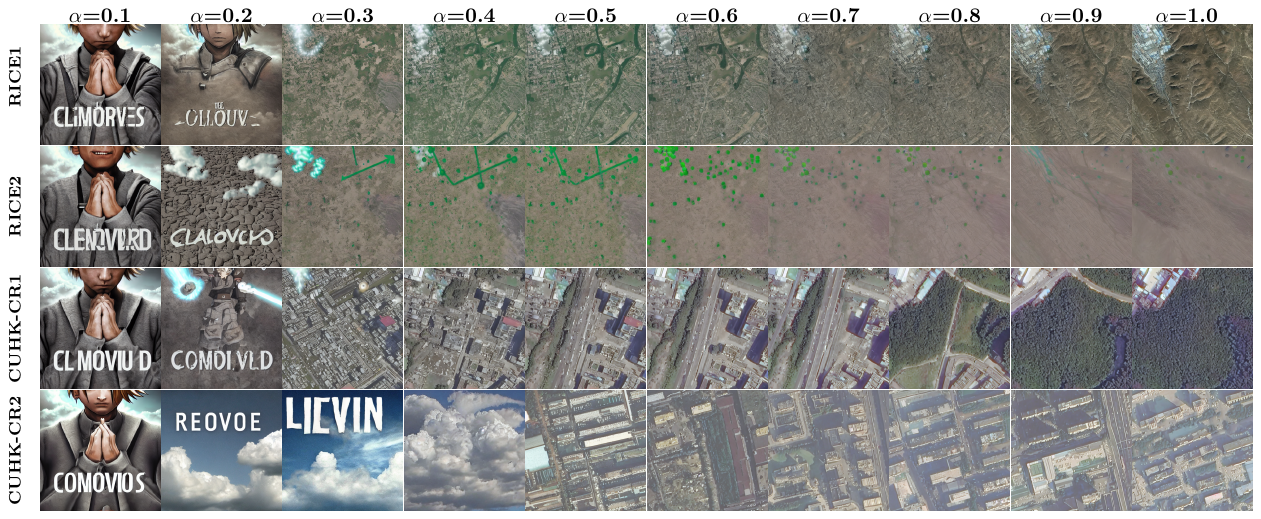} 
    \caption{Comparison of weighting scalling factor \(\alpha\) settings. It demonstrates that \(\alpha \geq 0.5\) lead to better outcomes.}
    \label{fig:lora_compare_imgs}
\end{figure*}

\begin{table*}[t]
    \caption{Ablation results of the DC module. It analyses the impact of integrating the DC module into various reference models.}
    \centering
    \resizebox{1.0\linewidth}{!}{
    \begin{tabular}{c|c|ccc|ccc|ccc|ccc}
    \hline
        \multirow{2}{*}{\textbf{Method}} & \multirow{2}{*}{\textbf{DC}} & \multicolumn{3}{|c|}{\textbf{RICE1}} & \multicolumn{3}{|c|}{\textbf{RICE2}} & \multicolumn{3}{|c|}{\textbf{CUHK-CR1}} & \multicolumn{3}{|c}{\textbf{CUHK-CR2}} \\ \cline{3-14}
        ~ & ~ & \textbf{PSNR}$\uparrow$ & \textbf{SSIM}$\uparrow$ & \textbf{LPIPS}$\downarrow$ & \textbf{PSNR}$\uparrow$ & \textbf{SSIM}$\uparrow$ & \textbf{LPIPS}$\downarrow$ & \textbf{PSNR}$\uparrow$ & \textbf{SSIM}$\uparrow$ & \textbf{LPIPS}$\downarrow$ & \textbf{PSNR}$\uparrow$ & \textbf{SSIM}$\uparrow$ & \textbf{LPIPS}$\downarrow$ \\ \midrule
        - & \checkmark & 35.542  & 0.9887  & 0.0054  & 33.678  & 0.8859  & 0.0451  & 26.291  & 0.7790  & 0.0285  & 24.595  & 0.6986  & 0.0326  \\ \cline{1-14}
        \multirow{2}{*}{MemoryNet} & $\times$ & 34.427 & 0.9865 & 0.0067 & 32.889 & 0.8801 & 0.0557 & 26.073 & 0.7741 & 0.0315  & 24.224  & 0.6838  & 0.0403 \\ 
        ~ & \checkmark & \textbf{35.715}  & \textbf{0.9928}  & \textbf{0.0050}  & \textbf{33.796}  & \underline{0.8861}  & \underline{0.0450}  & \textbf{26.633}  & \textbf{0.8146}  & \textbf{0.0252}  & \textbf{24.805}  & \textbf{0.7134}  & \textbf{0.0302}  \\ \cline{1-14}
        \multirow{2}{*}{MSDA} & $\times$ & 34.569 & 0.9871 & 0.0073 & 33.166 & 0.8793 & 0.0463 & 25.435 & 0.7483 & 0.0374  & 23.755  & 0.6661  & 0.0433 \\ 
        ~ & \checkmark & \underline{35.611}  & \underline{0.9893}  & \underline{0.0052}  & \underline{33.712}  & \textbf{0.8899}  & \textbf{0.0440}  & \underline{26.434}  & \underline{0.7908}  & \underline{0.0267}  & \underline{24.614}  & \underline{0.7043}  & \underline{0.0312}  \\ \bottomrule
    \end{tabular}
    }\label{table:ablation}
\end{table*}

\begin{table*}[ht]
    \caption{Weight scalling factor \(\alpha\) settings results. It indicates that \(\alpha = 0.7\) is the optimal setting, balancing performance and visual fidelity.}    
    \centering
    \resizebox{1.0\linewidth}{!}{
    \begin{tabular}{c|ccc|ccc|ccc|ccc}
    \toprule
        \multirow{2}{*}{\textbf{$\alpha$}} & \multicolumn{3}{|c|}{\textbf{RICE1}} & \multicolumn{3}{|c|}{\textbf{RICE2}} & \multicolumn{3}{|c|}{\textbf{CUHK-CR1}} & \multicolumn{3}{|c}{\textbf{CUHK-CR2}} \\ \cline{2-13}
        ~ & \textbf{PSNR}$\uparrow$ & \textbf{SSIM}$\uparrow$ & \textbf{LPIPS}$\downarrow$ & \textbf{PSNR}$\uparrow$ & \textbf{SSIM}$\uparrow$ & \textbf{LPIPS}$\downarrow$ & \textbf{PSNR}$\uparrow$ & \textbf{SSIM}$\uparrow$ & \textbf{LPIPS}$\downarrow$ & \textbf{PSNR}$\uparrow$ & \textbf{SSIM}$\uparrow$ & \textbf{LPIPS}$\downarrow$ \\ \midrule
        0.1 & 21.340  & 0.5860  & 0.2123  & 18.378  & 0.4790  & 0.2960  & 13.337  & 0.4975  & 0.2490  & 12.282  & 0.4938  & 0.2531  \\ 
        0.3  & 25.934  & 0.7162  & 0.1408  & 22.980  & 0.6100  & 0.1732  & 19.181  & 0.5317  & 0.1453  & 19.869  & 0.5971  & 0.1430  \\ 
        0.5  & 33.437  & 0.9574  & 0.0096  & 31.125  & 0.7458  & 0.0654  & 25.257  & 0.6256  & 0.0380  & 23.588  & 0.6738  & 0.0392  \\ 
        0.6  & 34.390  & 0.9887  & 0.0073  & 32.771  & 0.8393  & 0.0510  & 25.872  & 0.7192  & 0.0285  & 23.987  & 0.6877  & 0.0336  \\ 
        0.7  & \textbf{35.542}  & \textbf{0.9887}  & \textbf{0.0054}  & \textbf{33.678}  & \textbf{0.8859}  & \textbf{0.0451}  & \textbf{26.291}  & \textbf{0.7790}  & \textbf{0.0285}  & \textbf{24.595}  & \textbf{0.6986}  & \textbf{0.0326}  \\ 
        0.8  & 35.435  & \underline{0.9845}  & 0.0061  & \underline{33.676}  & \underline{0.8858}  & \underline{0.0456}  & \underline{26.288}  & 0.7724  & \underline{0.0292}  & 24.592  & 0.6964  & 0.0333  \\ 
        0.9  & \underline{35.441}  & 0.9826  & 0.0063  & 33.674  & 0.8825  & 0.0458  & 26.285  & 0.7707  & 0.0293  & 24.586  & \underline{0.6985}  & \underline{0.0331}  \\ 
        1.0  & 35.229  & 0.9795  & \underline{0.0057}  & 33.669  & 0.8837  & 0.0451  & 26.286  & \underline{0.7787}  & 0.0294  & \underline{24.593}  & 0.6889  & 0.0335  \\ \bottomrule
    \end{tabular}
    }\label{table:lora_alpha}
\end{table*}

\begin{table*}[t]
    \caption{$k$-means clustering statistics of datasets. It summarizes the number of images in each group after applying $k$-means clustering. The result provides insights into the size of each cluster and help understand the contribution of different groups for training.}
    \centering
    \resizebox{1.0\linewidth}{!}{
    \begin{tabular}{c|ccc|ccc|ccc|ccc}
    \toprule
        \multirow{2}{*}{\textbf{Group}} & \multicolumn{3}{|c|}{\textbf{RICE1}} & \multicolumn{3}{|c|}{\textbf{RICE2}} &  \multicolumn{3}{|c|}{\textbf{CUHK-CR1}} & \multicolumn{3}{|c}{\textbf{CUHK-CR2}} \\ \cline{2-13}
        ~ & \textbf{Group-1} & \textbf{Group-2} & \textbf{Group-3} & \textbf{Group-1} & \textbf{Group-2} & \textbf{Group-3} & \textbf{Group-1} & \textbf{Group-2} & \textbf{Group-3} & \textbf{Group-1} & \textbf{Group-2} & \textbf{Group-3} \\ \midrule
        No grouping & 500 & - & - & 736 & - & - & 534 & - & - & 448 & - & - \\ 
        2 Groups & 371 & 129 & - & 229 & 507 & - & 178 & 356 & - & 330 & 118 & - \\ 
        3 Groups & 100 & 334 & 66 & 356 & 320 & 60 & 229 & 72 & 233 & 215 & 77 & 156 \\ \bottomrule
    \end{tabular}
    }\label{table:progressive_data}
\end{table*}

\begin{figure*}[!ht]
    \centering
    \begin{subfigure}[b]{0.31\textwidth}
        \centering
        \includegraphics[width=\textwidth]{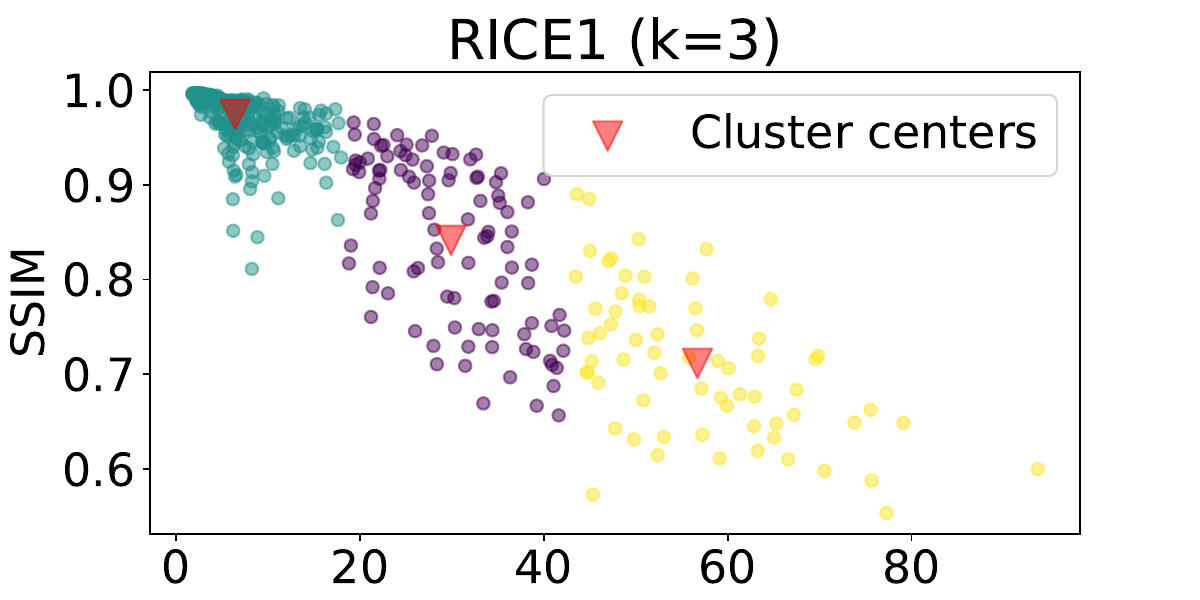}
        \caption{Grouped result ($k=3$)}
    \end{subfigure}
    \begin{subfigure}[b]{0.33\textwidth}
        \centering
        \includegraphics[width=\textwidth]{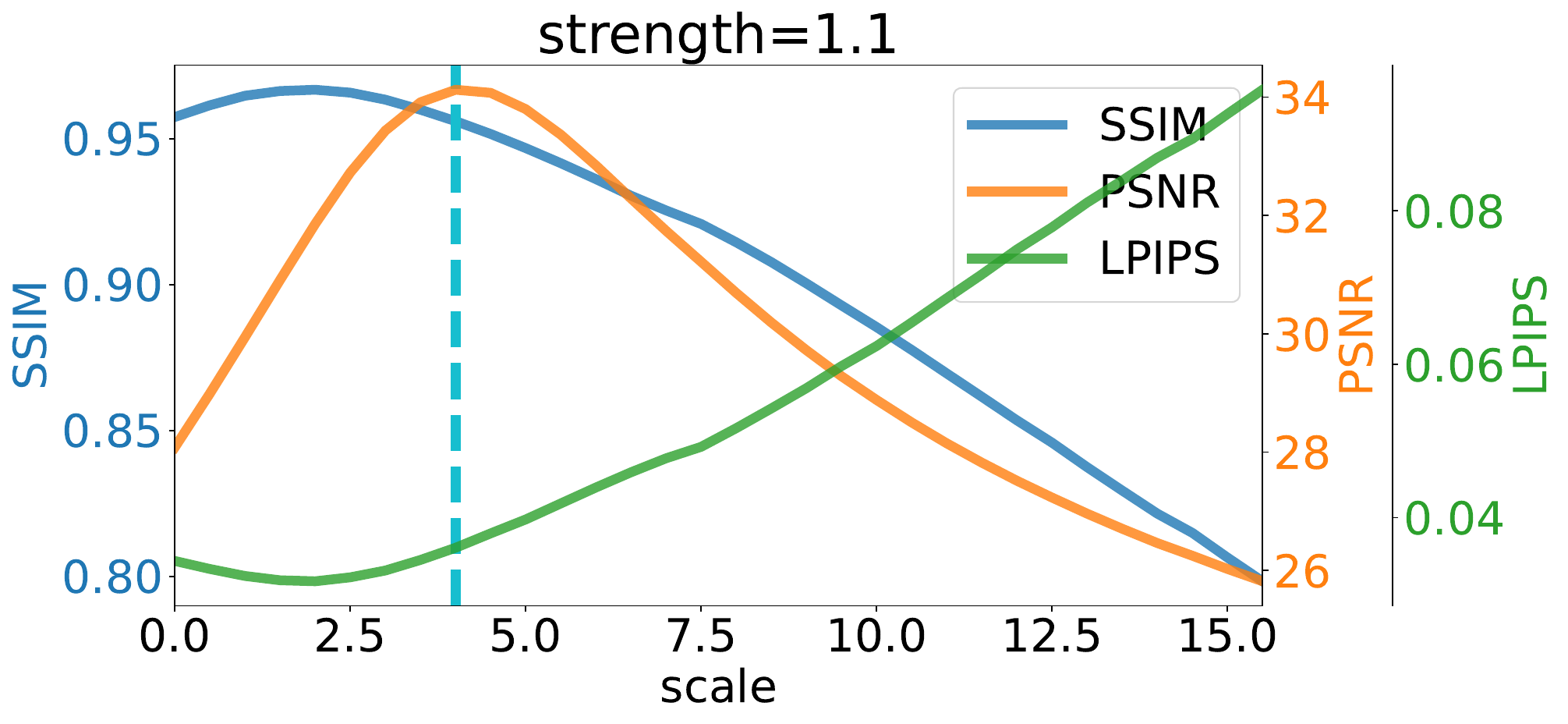}
        \caption{$scale$}
    \end{subfigure}
    \begin{subfigure}[b]{0.33\textwidth}
        \centering
        \includegraphics[width=\textwidth]{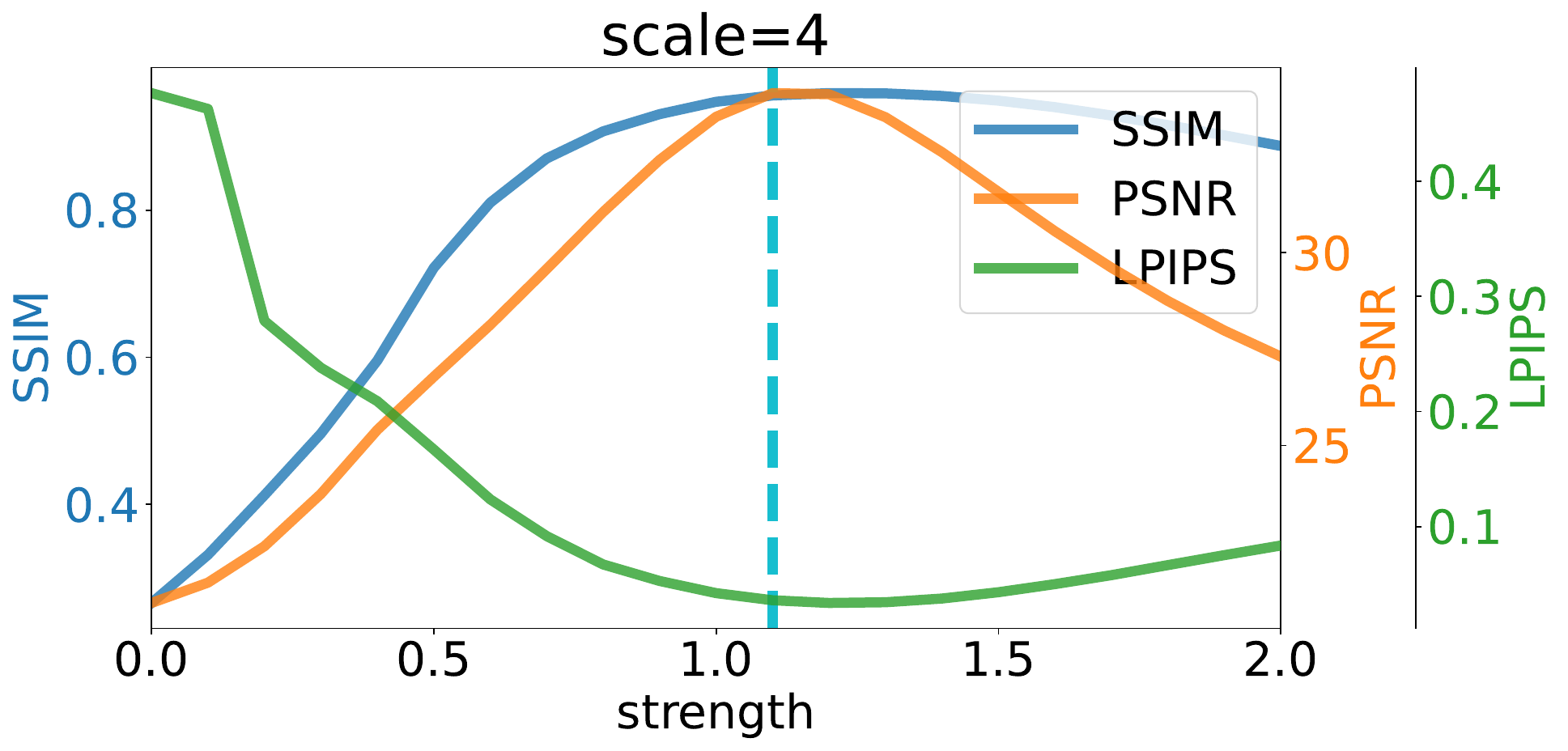}
        \caption{$strength$}
    \end{subfigure}    
    \caption{Grouped and controlling coefficients results. The optimal parameter values are $scale=4$ and $strength=1.1$.}
    \label{fig:params_controlnet}
\end{figure*}

\section{Experiment}
\subsection{Dataset Description}


To evaluate the efficiency of our proposed method, we utilize four datasets: RICE1/2 \citep{lin2019remote} and CUHK-CR1/2 \citep{sui2024diffusion}. The RICE dataset includes 500 images with thin cloud covers (RICE1) and 736 with thick cloud covers (RICE2). These images are randomly divided into an 80\% training set and a 20\% test set to ensure robust evaluation under various cloud conditions. Additionally, the CUHK-CR dataset comprises 1500 high-resolution images. The CUHK-CR dataset is further divided into CUHK-CR1 for thin clouds and CUHK-CR2 for thick clouds, facilitating a thorough assessment of our method's effectiveness and robustness under varied imaging conditions.


\subsection{Implementation Details}
All experiments were conducted on a single NVIDIA GeForce RTX 4090 GPU equipped with 24GB RAM, which provides high computational power and efficient processing capabilities, essential for handling large datasets and complex neural network operations. For the diffusion model generation stage, we used the pre-trained SD V1.5 with an instance prompt \texttt{remove <thin/thick> cloud}. 
The batch size was 4, and a learning rate of $1 \times 10^{-4}$ using a constant rate scheduler, with training capped at 30,000 steps. In the condition control stage, we also used a learning rate of $1 \times 10^{-4}$ without weight decay and a batch size of 4. The Adam optimizer ($\beta_1$=0.5, $\beta_2$=0.9) was applied over 50 epochs, with Mean Squared Error (MSE) as the loss function.



\subsection{Evaluation Metrics}

To evaluate the performance, we use three metrics: Peak Signal-to-Noise Ratio (PSNR) \citep{hore2010image}, Structural Similarity Index (SSIM) \citep{hore2010image}, and Learned Perceptual Image Patch Similarity (LPIPS) \citep{zhang2018unreasonable}. These metrics assess both technical accuracy and perceptual quality. \textbf{PSNR} measures pixel-level fidelity, with higher values indicating better image quality. \textbf{SSIM} assesses structural similarity, with values from -1 to 1; higher values indicate greater similarity. \textbf{LPIPS} evaluates perceptual similarity using deep learning features; lower scores indicate higher similarity.

\subsection{Comparison}

We compare our DC4CR model with SOTA cloud removal methods, including SpA-GAN \citep{pan2020cloud}, AMGAN-CR \citep{xu2022attention}, CVAE \citep{ding2022uncertainty}, MemoryNet \citep{zhang2023memory}, MSDA-CR \citep{yu2022cloud}, DE-MemoryNet \citep{sui2024diffusion}, and DE-MSDA \citep{sui2024diffusion}. The results are shown in Table \ref{table:comparision}, Figure \ref{fig:model_compare}, and Appendix.

As shown in Table \ref{table:comparision}, DC4CR achieves the highest PSNR (35.54 dB) and SSIM (0.9887) while obtaining the lowest LPIPS (0.0054) on RICE1, demonstrating superior image fidelity and perceptual quality. These improvements indicate better detail preservation and reduced artifacts compared to existing methods. 
Figure \ref{fig:model_compare} illustrates that DC4CR effectively removes clouds while maintaining textures and structural details. Compared to DE-MemoryNet and DE-MSDA, our model produces cleaner reconstructions, especially in complex scenes with vegetation and urban areas.

DC4CR eliminates the need for pre-generated cloud masks, learning directly from cloud-covered images. The integration of enhancement parameters further refines feature extraction, improving adaptability to varying cloud conditions. Additionally, its prompt-driven control allows flexible cloud removal settings, enhancing practicality in remote sensing applications.

\subsection{Ablation Study}
\textbf{Grouped Learning.}  
To evaluate the effectiveness of progressive learning, we applied $k$-means clustering ($k=2, 3$) to divide datasets into groups of increasing complexity. 
Appendix visualizes the clustering results, where samples are distributed based on SSIM and MSE. Table \ref{table:progressive_data} quantifies the dataset partitioning, showing that a three-group division yields the best balance. 
Table \ref{table:progressive} shows that the three-group setting consistently improves PSNR (by 0.5–1 dB) and SSIM while reducing LPIPS, especially in complex cloud cases. Without grouping, models struggle with diverse cloud conditions, confirming that progressive learning enhances stability and generalization.


\textbf{DC4CR Variants.}  
To assess the impact of our DC module, we integrated it into existing cloud removal models, specifically MemoryNet and MSDA. Figure \ref{fig:model_compare} illustrates qualitative improvements in cloud removal performance, showing sharper textures and fewer artifacts compared to baseline models. Table \ref{table:ablation} provides quantitative comparisons, where DC-MemoryNet achieves the highest PSNR and SSIM, outperforming DC-MSDA by 0.3 dB on average. Notably, on RICE2, DC-MemoryNet excels due to its superior handling of thick cloud occlusion, with a PSNR gain of 0.5 dB over DC-MSDA. These results validate the effectiveness of the DC module in enhancing cloud removal.


\subsection{Quantitative and Qualitative Analysis}

\textbf{Weight Scaling Factor. }
We investigated the effect of the weight scaling factor \(\alpha\) (ranging from 0 to 1) on image generation quality across different \(\alpha\) values. When \(\alpha < 0.3\), the model is largely ineffective, especially on the CUHK-CR2 dataset, where \(\alpha < 0.5\) shows almost no effectiveness (Figure \ref{fig:lora_compare_imgs}). At \(\alpha = 0.3\) and \(\alpha = 0.4\), image quality is poor, with noticeable artifacts and distortions. However, with \(\alpha \geq 0.5\), image quality begins to normalize, improving as \(\alpha\) increases, reaching peak performance at \(\alpha = 0.7\) (see Table \ref{table:lora_alpha}). Beyond \(\alpha \geq 0.5\), changes in evaluation metrics are minimal, indicating that \(\alpha\) directly influences image generation quality. While higher \(\alpha\) values yield clearer images, they may also distort land features, affecting the realism of the cloud removal effect. Therefore, selecting an appropriate \(\alpha\) is crucial for balancing quality and realism.

\textbf{Controlling Coefficients.}
The $scale$ parameter controls the prompt's influence on content generation, while $strength$ adjusts the intensity of conditioning input, affecting how closely the generated image matches the input conditions. Optimal settings of $scale=4$ and $strength=1.1$ (Figure \ref{fig:params_controlnet}) were identified, yielding the best image quality and accuracy. These findings highlight that proper parameter tuning significantly enhances model performance, aligning generated images more closely with expected outcomes. By adjusting $scale$ and $strength$, we observed improved results across various tasks, allowing for better customization and optimization to meet specific application needs.

\textbf{Enhancement Parameters. }
The parameter \(s\) adjusts feature amplitude through scaling, while \(b\) modifies feature translation via bias. Specifically, \(s_1\) and \(s_2\) scale features at different levels, and \(b_1\) and \(b_2\) adjust bias at specific levels. The parameters exhibit significant fluctuations based on the three evaluation metrics across the dataset (Figure Appendix.
On the RICE1, the optimal parameter combination, relatively speaking, is \(s_1 = 0.9\), \(s_2 = 0.4\), \(b_1 = 1.1\), and \(b_2 = 1.1\). Using this parameter set achieves relatively good image quality and accuracy. This indicates that a suitable parameter setting is crucial for enhancing the performance of model. By comparing different parameter combinations, we observe that adjusting the \(s\) and \(b\) parameters significantly influences the feature representation of generated images, leading to better results in various tasks. Although there are fluctuations in the results, choosing appropriate parameters can effectively improve the model's robustness and adaptability.

\section{Conclusion}

We propose \textbf{DC4CR}, a diffusion-based framework for cloud removal in remote sensing imagery. Our method eliminates the need for pre-generated cloud masks by leveraging {prompt-driven control}, enhancing preprocessing efficiency and adaptability. By integrating {precise generation control} and {enhancement mechanisms}, DC4CR achieves high-fidelity image restoration and state-of-the-art performance across multiple datasets. Additionally, our {modular design} allows seamless integration into existing cloud removal frameworks, further improving their effectiveness. These advancements establish DC4CR as a powerful and flexible solution for remote sensing image processing.






\bibliographystyle{named}
\bibliography{output}


\end{document}